\documentclass{article}

\usepackage{microtype}
\usepackage{graphicx}
\usepackage{subfigure}
\usepackage{booktabs} % for professional tables
\usepackage{dblfloatfix}

\usepackage{multirow}

\usepackage{hyperref}

\usepackage{scalerel}

\usepackage[accepted]{icml2024}

\usepackage{amsmath}
\usepackage{amssymb}
\usepackage{mathtools}
\usepackage{amsthm}

\usepackage[capitalize,noabbrev]{cleveref}

\theoremstyle{plain}

\theoremstyle{definition}

\theoremstyle{remark}

\usepackage[textsize=tiny]{todonotes}

\icmltitlerunning{Towards detailed and interpretable hybrid modeling of continental-scale bird migration}

\begin{document}

\twocolumn[
\icmltitle{Towards detailed and interpretable hybrid modeling \\ of continental-scale bird migration}




\begin{icmlauthorlist}
\icmlauthor{Fiona Lippert}{uva}
\icmlauthor{Bart Kranstauber}{uva}
\icmlauthor{Patrick Forr\'{e}}{uva}
\icmlauthor{E. Emiel van Loon}{uva}
\end{icmlauthorlist}

\icmlaffiliation{uva}{University of Amsterdam, Netherlands}

\icmlcorrespondingauthor{Fiona Lippert}{f.lippert@uva.nl}

\icmlkeywords{Machine Learning, ICML, hybrid modeling, ecology, forecasting, bird migration}

\vskip 0.3in
]



\printAffiliationsAndNotice{}  % leave blank if no need to mention equal contribution

\begin{abstract}

% General hybrid modeling
Hybrid modeling aims to augment traditional theory-driven models with machine learning components that learn unknown parameters, sub-models or correction terms from data. 
% FluxRGNN and bird migration
In this work, we build on FluxRGNN, a recently developed hybrid model of continental-scale bird migration, which combines a movement model inspired by fluid dynamics with recurrent neural networks that capture the complex decision-making processes of birds.
% Limitations of FluxRGNN
While FluxRGNN has been shown to successfully predict key migration patterns, its spatial resolution is constrained by the typically sparse observations obtained from weather radars. Additionally, its trainable components lack explicit incentives to adequately predict take-off and landing events. 
% Our contribution and results
Both aspects limit our ability to interpret model results ecologically. To address this, we propose two major modifications that allow for more detailed predictions on any desired tessellation while providing control over the interpretability of model components.
In experiments on the U.S. weather radar network, the enhanced model effectively leverages the underlying movement model, resulting in strong extrapolation capabilities to unobserved locations.
\end{abstract}

\section{Introduction}

Bird migration is a fascinating biological phenomenon with important implications for biodiversity and ecosystem functioning \cite{bauer2014migratory}. 
To effectively mitigate human-wildlife conflicts occurring during these seasonal mass movements \citep{loss2015direct, vandoren2017high, mclaren2018artificial}, near-term forecast models are essential tools. They enable stakeholders to anticipate peak migration events several hours to days in advance and prepare appropriate actions at the most critical times and locations \citep{vanbelle2007, dietze2018iterative, horton2021near, bradaric2024forecasting}.

Operational weather radar networks have become an invaluable data source for developing such forecasts. These radars can provide real-time information on bird densities and velocities across large geographical extents, making them particularly well suited for monitoring the long-distance migrations of nocturnally migrating songbirds \citep{dokter2011bird, bauer2017agricultural, nilsson2019revealing}.
However, the biological signals extracted from low-level radar scans typically provide only aggregated information about aerial movements within a relatively small area around each radar, leaving large gaps in-between \citep{dokter2011bird}.

To handle these spatially sparse observations, current continental-scale forecasting systems rely on purely data-driven models that link local environmental conditions such as wind, temperature and air pressure to observed bird densities \cite{vandoren2018continental}. 
While this approach is easy to implement in practice, it does not account for the underlying movement process and its spatio-temporal dependencies. 
This not only limits the accuracy of the forecast, but also prevents ecologists from gaining deeper insights into migration strategies, biomass flows, and connectivity.

On the other hand, recent studies have explored a fluid dynamics perspective on bird migration, where the spatio-temporal distribution of migrants is modeled with the continuity equation \citep{nussbaumer2021quantifying, nussbaumer2024turnover}. This allows information from weather radars to be integrated into a coherent description of biomass flows, capturing take-off, flight, and landing dynamics.
Building on this idea, \citet{lippert2022learning} have developed FluxRGNN, a hybrid forecast model that generates explicit predictions of bird fluxes across space and time while ensuring conservation of mass.

Hybrid or gray-box modeling aims to augment traditional theory-driven models with state-of-the-art machine learning \citep{willard2022integrating, yin2021augmenting, takeishi2021physics}. This approach has gained popularity especially in biology, ecology and the Earth sciences, where theoretical models are typically misspecified or only partially known \citep{senouf2023inferring, reichstein2019deep}.
In the case of FluxRGNN, the continuity equation is discretized using a finite volume scheme, which is then parameterized with recurrent neural networks that capture poorly understood decision making processes of birds and their complex dependencies on environmental conditions.
This results in an end-to-end differentiable recurrent graph neural network architecture that produces accurate and physically consistent forecasts while providing ecologically valuable insights into the migration process.

FluxRGNN handles spatially sparse radar observations by modeling migratory movements on the Voronoi tessellation of radar locations.
This, however, means that the forecast resolution is inherently restricted by the spatial distribution of radars.
In contrast, local data-driven models like \citet{vandoren2018continental} can, once trained, generate predictions for any location for which environmental predictors are available.

To bridge this gap, we propose an extension to FluxRGNN which decouples the underlying computational grid from the radar observation network and thereby allows for higher resolution forecasts on any desired tessellation.
In addition, we introduce an alternative flux parameterization which facilitates a straightforward integration with available velocity measurements, and therefore ensures meaningful estimates of implicitly learned take-off and landing processes.
Using data from the Next Generation Weather Radar (NEXRAD) network \citep{crum1993wsr, ansari2018unlocking}, which monitors bird migration across the contiguous United States, in combination with relevant environmental predictors we demonstrate the ability of our modified hybrid model to effectively balance forecast accuracy and interpretability, at both observed and unobserved locations.

\section{Problem setting}\label{sec:problem}

Consider a bounded spatial domain $\Omega\subset\mathbb{R}^2$ representing the air space above a geographic region of interest. 
We assume $\Omega$ to be partially observed by a set of $M$ radars $\mathcal{R}$. Each radar provides measurements $\mathbf{y}_m=\left[\rho_m, \mathbf{v}_m\right]$ of the vertically integrated bird density $\rho_m\in\mathbb{R}$ [birds/km\textsuperscript{2}] and average velocity $\mathbf{v}_m\in\mathbb{R}^2$ [km/h] %\textsuperscript{-1}] 
within a small area $A_m\subset\Omega$ around the radar location $\mathbf{x}_m\in\Omega$.
Given a sequence of radar measurements $\mathbf{Y}_{\mathcal{R}}^{(-\tau)}, \dots, \mathbf{Y}_{\mathcal{R}}^{(0)}\in\mathbb{R}^{3\times M}$ taken at time points $t_{-\tau}, \dots, t_0$, our goal is to predict the spatial distribution of migrating birds in $\Omega$ for $K$ future time points $t_1, \dots, t_K$.
In addition to the radar measurements, we assume access to relevant environmental variables for both past and future time points (e.g. based on operational weather forecasts) at the desired spatial resolution of the migration forecast.

\section{Background}\label{sec:background}

\subsection{Physical process model}

The highly synchronized mass movements of nocturnally migrating birds within a spatial domain $\Omega$ can be modeled with the continuity equation
\begin{align}\label{eq:continuity}
     \frac{\partial\rho}{\partial t} + \nabla \cdot (\rho\mathbf{v}) = s,
\end{align}
where $\rho$ is the density of birds in the air, $\mathbf{v}$ is the velocity field along which migrants fly across the continent and $s$ is a source/sink term capturing birds entering and leaving the sky \citep{nussbaumer2021quantifying, lippert2022learning}.
Importantly, both $\mathbf{v}$ and $s$ are the result of many migrants with different biological traits, strategies and past experiences making decisions based on dynamically changing environmental conditions.
As migratory movements may extend beyond the considered domain $\Omega$, Neumann boundary conditions
$\frac{\partial\rho}{\partial \mathbf{n}}(t, \mathbf{x}) = 0 \ \forall \mathbf{x}\in\partial\Omega,$
with $\mathbf{n}$ denoting the normal to $\partial\Omega$, can be used to allow for varying biomass flow into and out of $\Omega$.

\subsection{FluxRGNN}

FluxRNN \citep{lippert2022learning} combines this mechanistic movement model with the flexibility of deep neural networks to account for complex dependencies of bird behaviors on environmental conditions. As a result, it generates accurate and physically-consistent migration forecasts, while providing ecologically valuable insights into the spread and accumulation of migrants across space and time.
As FluxRGNN builds on the finite volume method (FVM) for spatial discretization, it ensures local mass conservation and is applicable to arbitrary (sparse and unstructured) observation networks.

\paragraph{Discretization}
More specifically, the spatial domain $\Omega$ is partitioned into $N$ finite volumes or cells $\mathcal{C}=\{C_i\}_{i=1}^N$, defined by the Voronoi tessellation of radar locations $\{\mathbf{x}_m\}_{m=1}^M$, for which the integral form of eq.~\ref{eq:continuity} is solved.
The dual of this tessellation is a graph $\mathcal{G}=(\mathcal{C}, \mathcal{E}_{\mathcal{C}})$ with $\mathcal{N}_{\mathcal{C}}(i)=\{C_j \ | \ (i,j)\in\mathcal{E}_{\mathcal{C}}\}$ denoting the local neighborhood of cell $C_i$. 
In combination with a forward Euler time discretization scheme, one obtains a system of algebraic equations
\begin{align}\label{eq:discrete_cell_fluxes}
     \scaleto{{\rho}_i^{(k+1)}\  =  \ {\rho}_i^{(k)} - \frac{1}{|C_i|}\sum_{j\in\mathcal{N}_{\mathcal{C}}(i)} F_{i\to j}^{(k\to k+1)} \ + \ {s}_i^{(k\to k+1)}}{26pt}
\end{align}
for all cells $C_i\in\mathcal{C}$. This defines an explicit update scheme for cell averages $\rho_i$ in terms of average source/sink terms $s_i$ and numerical flux terms $F_{i\to j}$ that capture the net movement from cell $C_i$ to a neighboring cell $C_j$.
Here, superscripts $(k)$ represent quantities at time point $t_k$, and superscripts $(k\to k+1)$ refer to terms that have been integrated over the time interval $\left[t_k, t_{k+1}\right]$.

\paragraph{Hybrid neural-numerical architecture}
Given knowledge about current bird densities ${\boldsymbol{\rho}}_{\mathcal{C}}^{(0)}\in\mathbb{R}^N$ across the tessellation $\mathcal{C}$, eq.~\ref{eq:discrete_cell_fluxes} can in principle be used to simulate the system forward in time and generate forecasts of future bird densities ${\boldsymbol{\rho}}^{(1)}_{\mathcal{C}}, \dots, {\boldsymbol{\rho}}^{(K)}_{\mathcal{C}}$.
In practice, however, the necessary flux and source/sink terms are unknown and their dependencies on dynamic environmental conditions only poorly understood.
Therefore, FluxRGNN employs a set of neural networks to learn mappings from environmental conditions and past radar observations to fluxes and source/sink terms.
This results in a hybrid neural-numerical architecture that inherits its graph structure from the FVM discretization.
As eq.~\ref{eq:discrete_cell_fluxes} is differentiable, the model can be trained end-to-end to minimize the mismatch between observed bird densities and the corresponding cell predictions. 

\paragraph{Encoder-decoder backbone}
The learnable fluxes and source/sink terms are the result of individual migrants continuously taking decisions based on past and current experiences. 
To adequately capture these complex temporal dependencies, FluxRGNN uses a recurrent encoder-decoder backbone, which iteratively integrates information about environmental conditions $\mathbf{U}^{(-\tau)},\dots,\mathbf{U}^{(K)}$, radar measurements $\mathbf{Y}^{(-\tau)}, \dots, \mathbf{Y}^{(0)}$, and predicted densities $\hat{\boldsymbol{\rho}}^{(0)},\dots,\hat{\boldsymbol{\rho}}^{(K)}$ into cell representations
\begin{align}
    \mathbf{Z}^{(k)} &= \text{RNN}_{\text{enc}}(\mathbf{Z}^{(k-1)}, \mathbf{Y}^{(k)}, \mathbf{U}^{(k)}, \mathbf{G}) & \forall k \leq 0, \label{eq:encoder} \\
    \mathbf{Z}^{(k)} &= \text{RNN}_{\text{dec}}(\mathbf{Z}^{(k-1)}, \hat{\boldsymbol{\rho}}^{(k-1)}, \mathbf{U}^{(k)}, \mathbf{G})  & \forall k > 0, \label{eq:decoder}
\end{align}
where $\mathbf{G}$ are static features of the tessellation $\mathcal{G}$.
These representations are then used to predict fluxes and source/sink terms. %(see appendix for more details).

\section{Improved hybrid modeling}\label{sec:method}

In the following, we introduce two major modifications to the FluxRGNN framework: (i) an extension allowing for migration forecasts on arbitrary tessellations, and (ii) an adjusted neural flux parameterization which, in combination with additional supervision of predicted movements, facilitates more reliable insights into the migration process.

\subsection{From Voronoi cells to arbitrary tessellations}
\label{sec:method:tessellation}

The original FluxRGNN framework \citep{lippert2022learning} models movements on the Voronoi tessellation of radar locations. 
While this ensures a one-to-one mapping between observations and grid cells, making it straight forward to process radar data in the encoder (eq.~\ref{eq:encoder}) and to initialize cell predictions at $t_0$, it also significantly limits the spatial resolution of predictions and may lead to artifacts due to highly irregular cell shapes and sizes.
To facilitate more detailed and robust forecasts on any desired grid, we propose an extension which makes the definition of $\mathcal{C}$ independent of $\mathcal{R}$.
This requires adjustments to the encoder, initial state and comparison to radar observations, establishing connections between cell space and measurement space.
The resulting model operates on a graph $(\mathcal{C}, \mathcal{R}, \mathcal{E}_{\mathcal{C}}, \mathcal{E}_{\mathcal{R}\to\mathcal{C}}, \mathcal{E}_{\mathcal{C}\to\mathcal{R}})$ with cell-to-cell edges $\mathcal{E}_{\mathcal{C}}$, radar-to-cell edges $\mathcal{E}_{\mathcal{R}\to\mathcal{C}}$, and cell-to-radar edges $\mathcal{E}_{\mathcal{C}\to\mathcal{R}}$.

\paragraph{Notation}
In the remainder of this paper, we distinguish cell-level quantities from radar-level quantities using subscripts $\mathcal{C}$ and $\mathcal{R}$, respectively.

\begin{figure}[htb]
    \centering
    \includegraphics[width=0.45\textwidth]{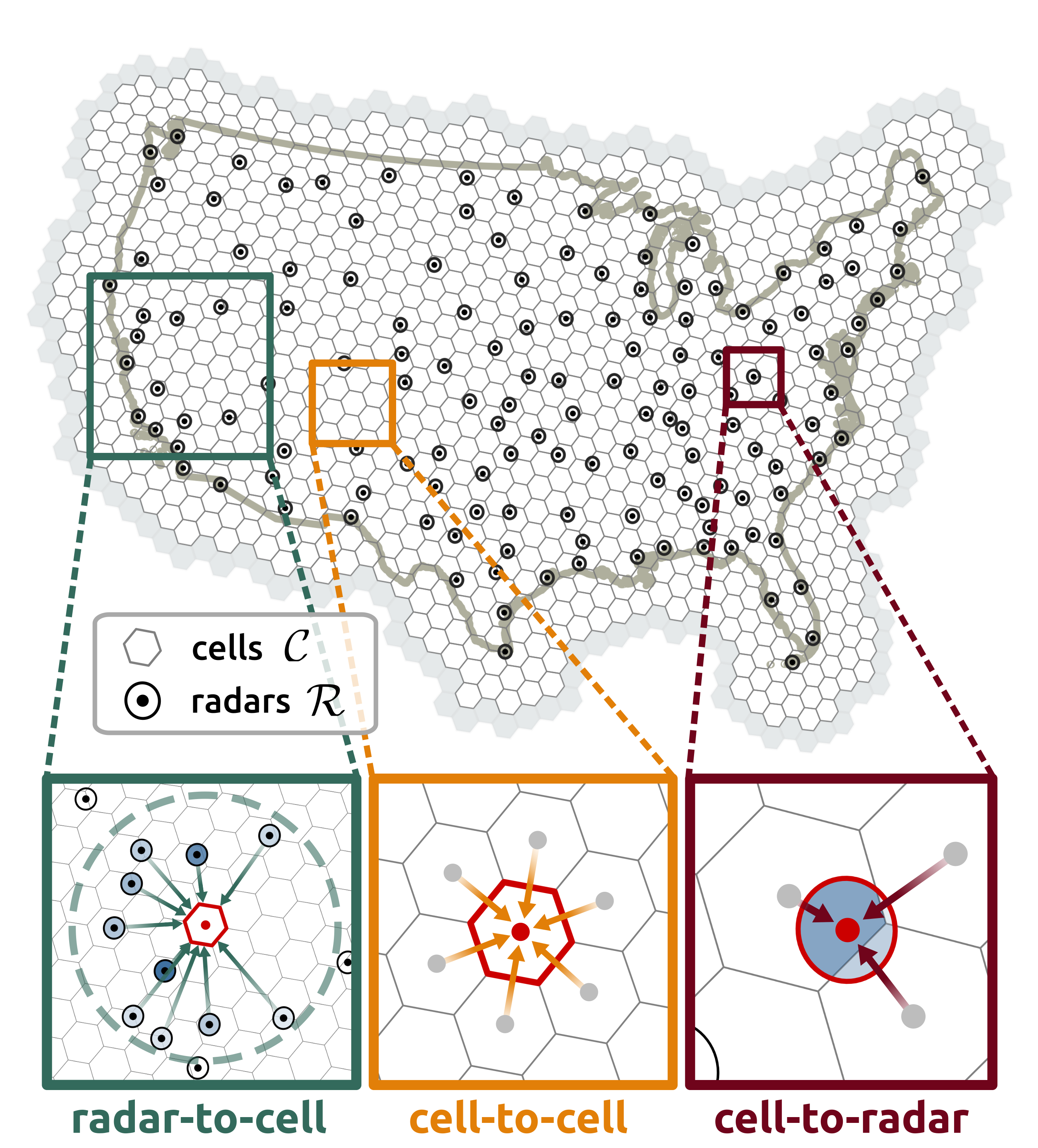}
    \caption{Map of the NEXRAD weather radar network, with black circles indicating measurement areas $A_m$, and the hexagonal tessellation on which movements are modeled. During encoding, $f_{\mathcal{R}\to\mathcal{C}}$ maps sparse radar measurements to cell space (bottom left), in which the forecast is generated based on within-cell source/sink terms $s_i$ and cell-to-cell fluxes $F_{j\to i}$ (bottom center). Finally, $f_{\mathcal{C}\to\mathcal{R}}$ maps cell-level predictions back to measurement space (bottom right).}
    \label{fig:radar_and_cell_overview}
\end{figure}

\paragraph{Radar-to-cell mapping}
 
To incorporate sparse radar measurements into cell-level encoder representations, we introduce a radar-to-cell mapping $f_{\mathcal{R}\to\mathcal{C}}$, which transforms each $\mathbf{Y}^{(k)}_{\mathcal{R}}\in\mathbb{R}^{3\times M}$ into pseudo-measurements $\Tilde{\mathbf{Y}}^{(k)}_{\mathcal{C}}\in\mathbb{R}^{H_Y\times N}$.
Note that this operation on $\mathcal{E}_{\mathcal{R}\to\mathcal{C}}$ can be anything from a simple spatial interpolation to a graph neural network layer.
The cell-level pseudo-measurements are then fed to the encoder as before (see eq.~\ref{eq:encoder}).
After encoding all data up to time $t_0$, we predict the initial cell states as
${\boldsymbol{\rho}}_{\mathcal{C}}^{(0)}=f_{\text{initial}}(\Tilde{\mathbf{Y}}^{(0)}_{\mathcal{C}}, \mathbf{Z}^{(0)}_{\mathcal{C}})$.
Again, $f_{\text{initial}}$ can be anything from a copy of interpolated radar measurements to a graph neural network layer.

\paragraph{Cell-to-radar mapping}

During model training and evaluation, cell-level predictions need to be compared to ground truth radar measurements.
For this purpose, we introduce a cell-to-radar mapping or observation model $f_{\mathcal{C}\to\mathcal{R}}$, which transforms cell-level quantities into radar-level quantities by taking a weighted average on $\mathcal{E}_{\mathcal{C}\to\mathcal{R}}$, where $(n, m)\in \mathcal{E}_{\mathcal{C}\to\mathcal{R}}$ if the measurement area $A_m$ overlaps with cell $C_n$. The corresponding weight is proportional to the area of the overlap, simulating uniform observations across $A_m$.

\paragraph{Learning to predict long sequences}
To guide the model in learning to make accurate predictions at large forecasting horizons without severe amplification of errors over time, 
\citet{lippert2022learning} use scheduled sampling \citep{bengio2015scheduled} with an exponentially decaying teacher forcing rate. With $\mathcal{C}$ being independent of $\mathcal{R}$, however, cell-level predictions cannot simply be replaced by radar observations to stabilize training. 
Instead, we gradually increase the forecasting horizon during training, similarly to \cite{lam2023learning}.
This allows the model to initially focus on learning basic short-term dynamics and only later on improve long-term stability.

\subsection{Balancing accuracy and interpretability}
\label{sec:methods:interpretability}

To accommodate settings where no reliable velocity measurements are available, the original FluxRGNN framework learns both spatial fluxes and local source/sink terms implicitly, only through supervision of bird densities. % in an unsupervised way.
While \citet{lippert2022learning} find that the inferred quantities correlate well with the underlying ground truth in a simulated setting, 
this unsupervised setup bears the risk of ignoring spatial dependencies caused by aerial movements and instead explaining all dynamics via local source/sink terms.

One way to avoid severe over-estimation of take-off and landing events would be to follow other hybrid modeling frameworks \citep{yin2021augmenting, takeishi2021physics} and interpret the source/sink term as a learnable correction term which should be minimized so that spatial fluxes explain as much of the data as possible.
However, take-off and landing are integral parts of the migration process. Such an approach would thus lead to severe under-estimation of mass migration events, which in turn limits the practical use of the resulting forecast model.

Instead, we exploit available velocity measurements for additional supervision.
For this purpose, we introduce a modified flux model that relies on explicit predictions of average cell velocities.
These velocities can than be included in the optimization objective to encourage agreement with observed movement patterns.

\paragraph{Flux parameterization}
Originally, FluxRGNN parameterizes the numerical flux terms $F_{i\to j}$ in terms of the proportions of birds moving between adjacent cells \cite{lippert2022learning}, circumventing the need for approximate FVM schemes.
To facilitate velocity-based supervision, we instead propose a neural parameterization of velocities $\mathbf{v}_i$, which are used to compute numerical flux terms
\begin{align}
    F_{i\to j}^{(k\to k+1)} &= F_{\text{FVM}}(\rho_i^{(k)}, \rho_j^{(k)}, \mathbf{v}_i^{(k)}, \mathbf{v}_j^{(k)}, \mathbf{G}),
\end{align}
where $F_{\text{FVM}}$ can be any numerical FVM scheme that approximates fluxes based on cell averages. For more details, see Appendix \ref{ap:models}.

\paragraph{Loss function}
Using the same observation model $f_{\mathcal{C}\to\mathcal{R}}$ for both bird densities and velocities, we extend the FluxRGNN loss function by a weighted velocity-based loss term.
The overall loss of a forecast up to time $t_K$ becomes
\begin{align}
\begin{split}
    \mathcal{L}_{1:K} = \frac{1}{K} & \sum_{k=1}^{K} \bigg[ \ \mathcal{L}_{\rho}\left( \boldsymbol{\rho}^{(k)}_{\mathcal{R}}, f_{\mathcal{C}\to\mathcal{R}}\left(\boldsymbol{\rho}^{(k)}_{\mathcal{C}}\right)\right) \\ 
    &+ \lambda \cdot \mathcal{L}_{v}\left( \mathbf{v}^{(k)}_{\mathcal{R}}, f_{\mathcal{C}\to\mathcal{R}}\left(\mathbf{v}^{(k)}_{\mathcal{C}}\right)\right)\bigg],
\end{split}
\end{align}
where $\lambda\geq 0$ determines the trade-off between accurate bird density predictions and reliable estimates of aerial movements and take-off/landing events.
The value of $\lambda$ should be chosen carefully depending on the model purpose and the faithfulness of the respective radar measurements.

\section{Experiments and results}\label{sec:results}

We apply our modified hybrid model (FluxRGNN+) to data from the Next Generation Weather Radar (NEXRAD) network \citep{crum1993wsr, ansari2018unlocking}, consisting of 143 radars distributed across the continental United States.
To capture the effects of environmental conditions on migration behaviors, we consider a range of atmospheric variables extracted from the ERA5 reanalysis dataset \citep{hersbach2020era5}. In addition, we use NLCD landcover data \citep{yang2018new} to include information on habitat types and other landscape characteristic.

\subsection{Experimental setup}
We restrict our experiments to the autumn migration season, between 1 August and 15 November.
To avoid information leakage between training and test datasets, we split our data by entire years, using 2013-2018 for model training, 2019 for hyperparameter tuning and model selection, and 2020-2021 for final model evaluation.
Each dataset is then split into 94h-sequences, where the first 24h are fed to the encoder as context information (if applicable), while the remaining hours are used to determine model performance. During training we use a maximum forecast horizon of $K=48$h, which we increase to $K=72$h during testing.
% During training, we use overlapping sequences starting at all possible hours of the day. For the final evaluation, we restrict this to sequences starting at 13:00 Eastern Time.

\begin{figure*}[!b]
    \centering
    \includegraphics[width=\textwidth]{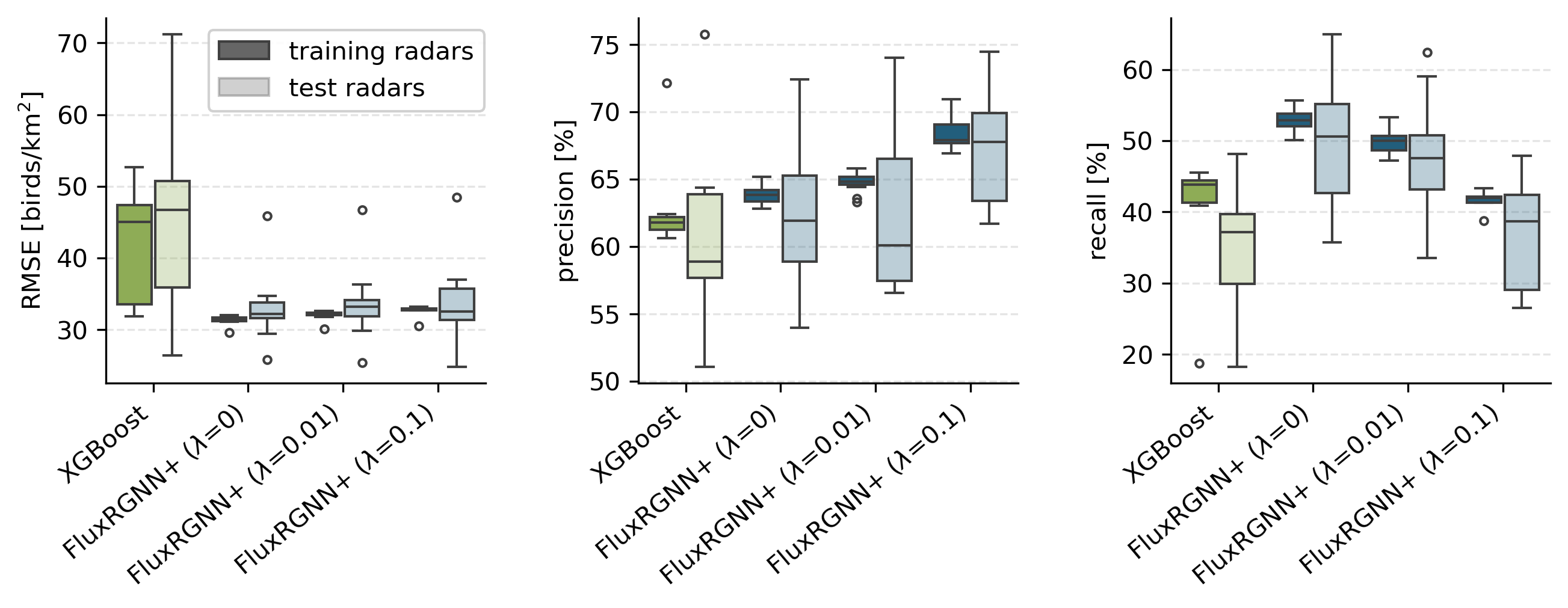}
    \vspace{-5ex}
    \caption{Spatial cross-validation of bird density predictions. Box plots show the variability of evaluation metrics across 10 cross-validation folds, where different subsets of radars were held out during training. Note that the subsets of training radars for different folds overlap, which naturally leads to less variability in evaluation metrics than for the distinct subsets of test radars.}
    \label{fig:cv_performance}
\end{figure*}
\begin{figure*}[htb]
    \centering
    \includegraphics[width=\textwidth]{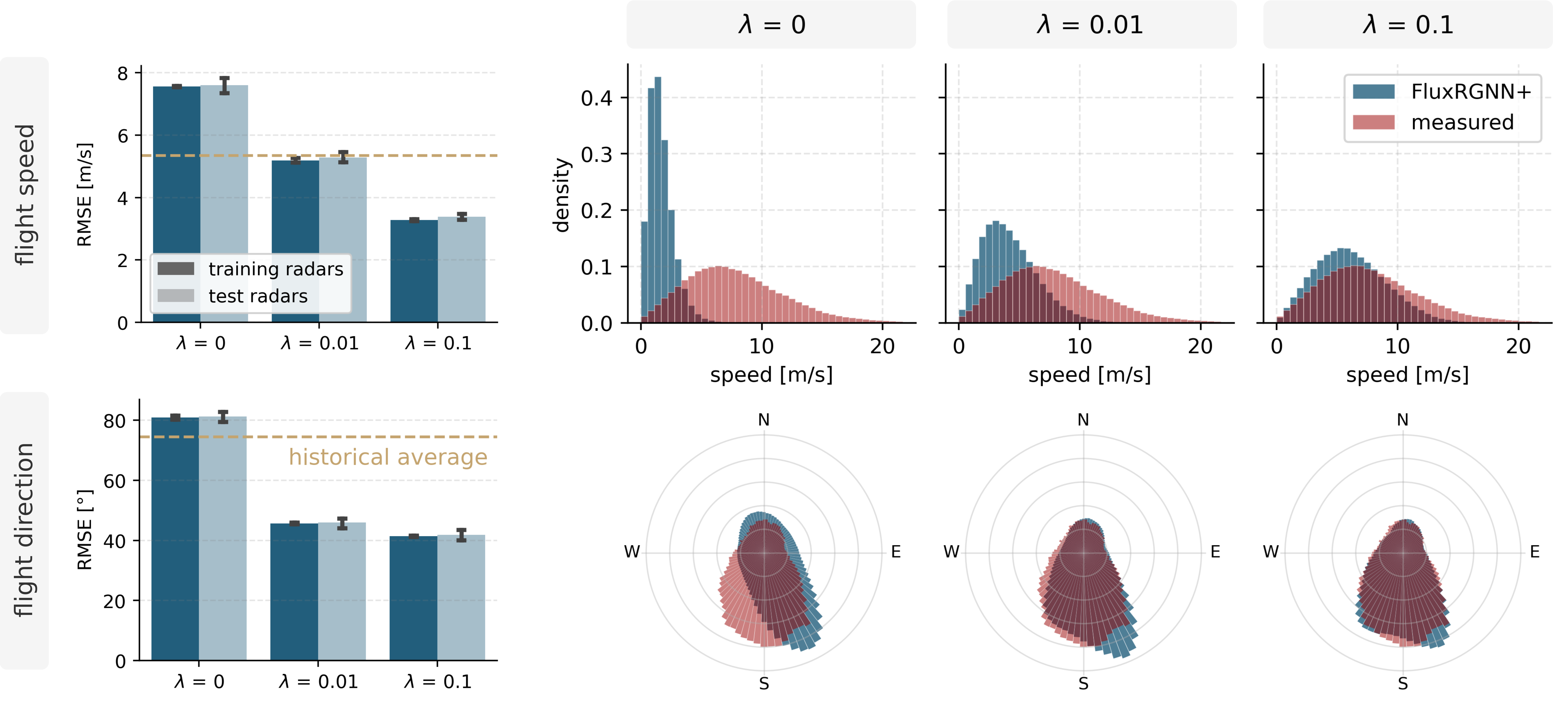}
    \vspace{-5ex}
    \caption{Evaluation of flight speeds and directions predicted by FluxRGNN+ trained with varying $\lambda$. Left: Results of the spatial cross-validation for training and test radars respectively. Right: Histograms of predicted and measured quantities for held out test radars across all 10 cross-validation folds.}
    \label{fig:speed_dir_performance}
\end{figure*}

\subsection{FluxRGNN+ setup}

\paragraph{Tessellation} The FluxRGNN+ extension allows us to model migratory movements on any desired tessellation. 
Here, we use a regular grid based on Uber's Hexagonal Hierarchical Spatial Index \citep{H3}, a standardized global tessellations consisting of approximately equally sized hexagons, which is particularly well suited for movement modeling \citep{birch2007rectangular}.
The resulting tessellation consists of 1091 cells with diameters ranging between 125 and 150km. 
At this resolution, regional variation due to urban areas, coastlines or mountain ranges can be (partially) resolved. At the same time, cells remain large enough to model movements at an hourly resolution without violating the continuity assumption, which reduces computational costs compared to finer temporal resolutions.

\paragraph{Radar-to-cell mapping}
To ensure that all cells receive information on radar measurements, we define $\mathcal{E}_{\mathcal{R}\to\mathcal{C}}$ based on the $k$-nearest-neighbor graph connecting each cell $C_n$ to the $k=10$ closest radars. Cell-level quantities are then computed as weighted averages
\begin{align}
    \Tilde{y}_n^{(k)} = \sum_{(m, n)\in\mathcal{E}_{\mathcal{R}\to\mathcal{C}}} w_{n,m} \cdot y_m^{(k)}
\end{align}
where weights $w_{n, m}$ scale inversely with the distance $d(\mathbf{x}_n, \mathbf{x}_m)$ between the cell center and the respective radar location.
The same interpolation scheme is used to define initial cell states $\boldsymbol{\rho}_{\mathcal{C}}^{(0)}$.

\paragraph{Flux computation}
For ease of implementation, we use a simple upwind scheme to compute flux terms based on predicted cell densities and velocities. For more details, see Appendix \ref{ap:models}.

\subsection{Predictive performance}

Modeling movements on a regular hexagonal grid instead of the coarse Voronoi tessellation of radar locations has the potential to reduce artifacts related to irregular cell shapes and sizes.
At the same time, the required radar-to-cell mapping may introduce interpolation errors and result in information loss.
To investigate this trade-off, we compare the predictive performance of FluxRGNN+ to the original FluxRGNN, as well as a local regression model (XGBoost) similar to \citet{vandoren2018continental}, and a radar-specific historical average capturing typical daily and seasonal patterns.
To ensure a fair comparison, all models are trained only on bird densities, i.e. using $\lambda=0$.

We evaluate bird density predictions in terms of the RMSE across all radars and forecast horizons (up to 72h), excluding daytime hours.
In addition, we compute precision and recall based on a threshold of 150 birds/km\textsuperscript{2} (ca. 95\% quantile) to determine how well a model identifies rare high intensity migration events, which are of particular interest for conflict mitigation.

\begin{table}[htb]
    \caption{Evaluation of bird density predictions. All metrics are reported as mean $\pm$ std across 5 different random seeds.  \newline}
    \label{tab:all_radars_performance}
    \centering
    \scriptsize
    \begin{tabular}{lccc}
        \toprule
         & RMSE $\left[\text{birds/km}^2\right] \downarrow$ & precision $\left[\%\right] \uparrow$ & recall $\left[\%\right] \uparrow$ \\
        \midrule
        historical avg & $63.77 \pm 0.00$ & $32.05 \pm 0.00$ & $18.79 \pm 0.00$ \\
        XGBoost & $47.25 \pm 0.06$ & $61.18 \pm 0.14$ & $43.92 \pm 0.14$ \\
        FluxRGNN & $31.49 \pm 0.67$ & $62.59 \pm 3.04$ & ${54.90 \pm 4.84}$ \\
        \midrule
        FluxRGNN+ & ${31.64 \pm 0.40}$ & ${63.94 \pm 3.27}$ & ${53.11 \pm 5.69}$ \\
        \bottomrule
    \end{tabular}
\end{table}

In line with previous results from \citet{lippert2022learning}, we find that FluxRGNN generates more accurate bird density predictions than both the historical average and the local XGBoost model (see Table~\ref{tab:all_radars_performance}). In particular, we observe a substantial improvement in terms of recall, i.e. the percentage of correctly identified peak migration hours, which is of particular importance for conflict mitigation.
The performance of our modified FluxRGNN+ model is on par with the original FluxRGNN across all evaluation metrics, indicating that the reduction of tessellation-related artifacts compensates for any information loss due to radar-to-cell interpolation errors.
Importantly, this means that we are able to generate more informative, higher resolution forecasts without losing predictive power.

\subsection{Extrapolation to unobserved locations}\label{sec:results:extrapolation}

By extending FluxRGNN to arbitrary tessellations, we aim to resolve more local variability in migratory movements, for example in response to fine-scale weather patterns and habitat types.
To assess the quality of predictions at unobserved locations in-between radars, we perform a spatial cross-validation where we divide all available radars into 10 random subsets and train all models on 9 out of 10 of these subsets, leaving the rest for independent evaluation. Importantly, radar measurements of test radars are only used for evaluation, and not for providing context information to the encoder.
Repeating this process for all subsets gives an estimate of how well our model generalizes to unobserved locations.
As both the original FluxRNN and the historical average are tied to available radar measurements, we do not include them in this analysis.

Figure \ref{fig:cv_performance} summarizes the results of this spatial cross-validation. As expected, predictions for independent test radars are on average less accurate than for training radars. Nevertheless, FluxRGNN+ continues to perform better than XGBoost across all evaluation metrics.
A radar-specific evaluation reveals that for FluxRGNN+ ($\lambda=0$) the RMSE for an excluded radar increases on average by $9.67\%$ compared to a setting where this radar is included during training. For XGBoost, this value is almost three times as large ($26.45\%$), suggesting that even with $\lambda=0$, i.e. no supervision on velocities, the underlying movement model helps to make spatially consistent predictions.

\begin{figure*}[htb]
    \centering
    \includegraphics[width=\textwidth]{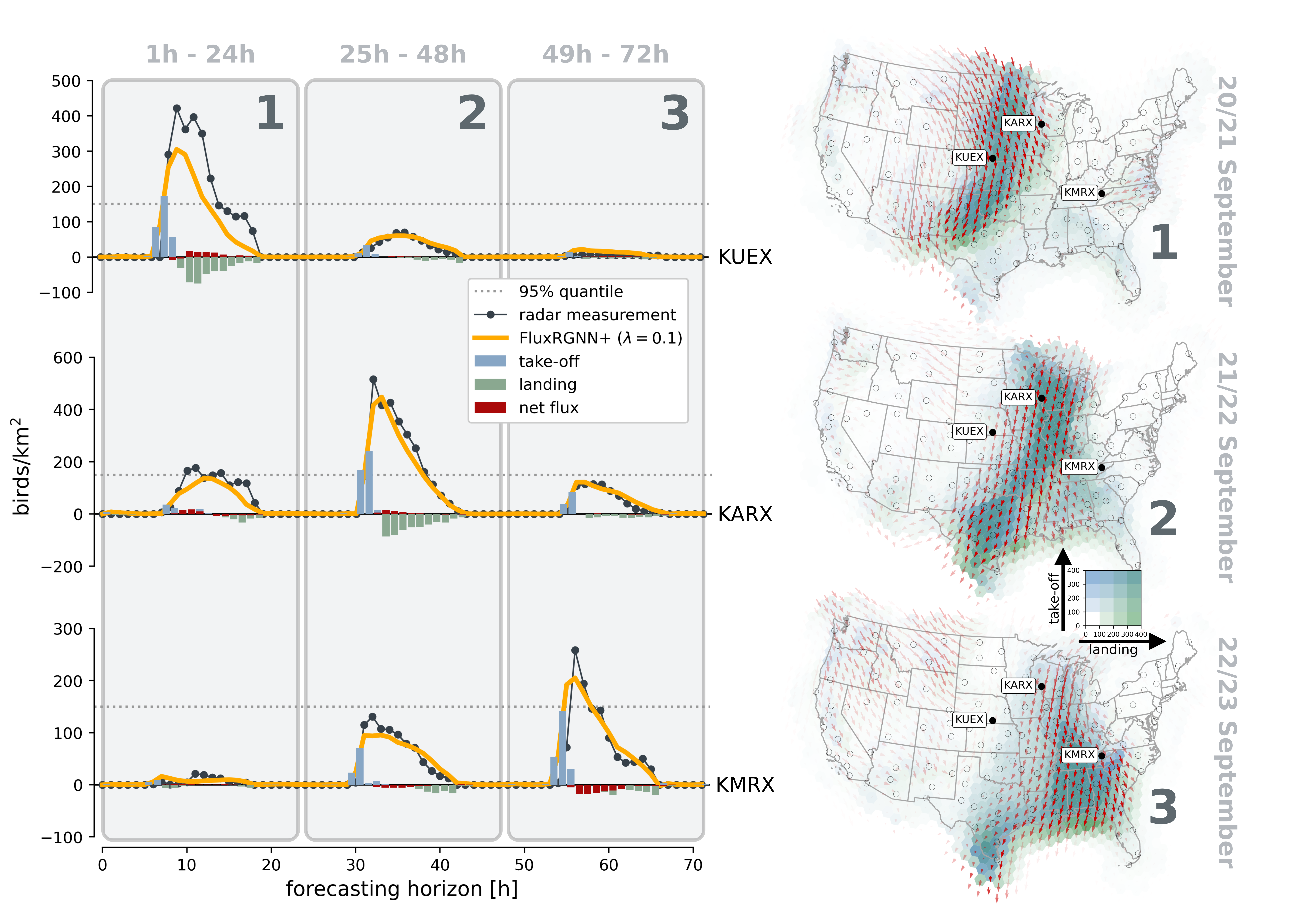}
    \caption{An example forecast of three consecutive high intensity migration nights (numbers 1-3) in September 2021 generated by FluxRGNN+ trained with $\lambda=0.1$ on years 2013-2020. The three time series on the left correspond to the radars marked in the maps on the right. To distinguish between take-off and landing, we separate hours with positive and negative source/sink term and aggregate them respectively. Red arrows on the maps indicate average velocities in areas with substantial migration.}
    \label{fig:example_forecast}
\end{figure*}

\subsection{Improving interpretability}

To investigate the trade-off between forecast accuracy and interpretability encoded by $\lambda$ (see Section~\ref{sec:methods:interpretability}), we gradually increase $\lambda$ and evaluate the effect on predicted bird densities, flight speed and directions, using the same spatial cross-validation setup as before.
To avoid issues with unreliable velocity measurements when hardly any birds are in the air, we only include moments with $\rho_m^{(k)} > 5 \ \text{birds/km}^2$ in our evaluation of velocity predictions and when computing historical averages.

As expected, increasing $\lambda$ results in predicted and observed flight speeds and directions matching more closely (see Figure \ref{fig:speed_dir_performance}). Notably, already a rather small contribution to the overall loss yields movement predictions that are more accurate than a simple historical average, which is a relatively strong baseline due to consistent continental-scale migration routes.
Overall, flight speeds seem to be more challenging to predict correctly, especially during peak migration where birds move much faster due to favorable wind conditions. However, with increased supervision ($\lambda=0.1$) the predictions of these high speed movements improve considerably.

At the same time, we observe a gradual degradation of bird density predictions in terms of RMSE and recall, indicating that forecasts of mass departure events become more conservative as the model focus is shifted towards accurate velocity predictions. This is supported by the accompanied increase in precision.
Nevertheless, the decrease in performance is rather small compared to the respective gain in terms of flight speed and directions. Even for the highest considered $\lambda$, bird density predictions remain more accurate or at least comparable to XGBoost. 
Finally, when considering the relative increase in RMSE when excluding a radar during training (see Section \ref{sec:results:extrapolation}), we find that increasing $\lambda$ actually improves the extrapolation capabilities of the model ($9.45\%$ for $\lambda=0.01$ and $7.69\%$ for $\lambda=0.1$).
This is likely due to FluxRGNN+ tending to under-predict flight speeds for small $\lambda$ (see Figure \ref{fig:speed_dir_performance}), resulting in very little information flow between distant locations. In contrast, as $\lambda$ increases aerial fluxes increase as well, which helps the model to exploit information from surrounding radars to make more accurate predictions at unobserved locations.

For applications where reliable extrapolation and interpretability of predicted movements are key, we thus recommend a setting of $\lambda=0.1$. Figure \ref{fig:example_forecast} shows an example forecast of three consecutive high intensity migration nights generated with this setting. The migration wave passing through the East of the continent, with distinct density peaks at radars falling within this wave, is clearly captured by the model.
Take-off and landing areas overlap substantially, confirming previous finding by \citet{nussbaumer2024turnover}.
% TODO: discuss this in more detail? E.g. overlap of take-off and landing with reference to Raphael?

\section{Discussion and conclusion}

We have presented FluxRGNN+, an extension to a recently developed hybrid model of continental-scale bird migration which allows us to make more detailed and interpretable bird migration forecasts on any desired tessellation.
The resulting bird density predictions at observed radar locations are on par with those generated by the original FluxRGNN model. At the same time, the underlying movement model enables FluxRGNN+ to make more accurate predictions at unobserved locations in-between radars than a purely local XGBoost baseline.
Increasing supervision on predicted bird velocities naturally yields more realistic movement patterns, which in turn facilitates even more robust extrapolation to unobserved locations. 

Nevertheless, there exists an inevitable trade-off between accurate forecasts of peak migration and reliable disentanglement of the underlying processes. The balance between the two objectives depends on the specific purpose of the forecast.
For example, for effective mitigation of collisions with wind turbines, aircraft or illuminated buildings, a focus on peak migration events is key \citep{bradaric2024forecasting, kranstauber2022ensemble, horton2021near, burt2023effects}.
Conversely, reliable estimates of aerial fluxes, take-off and landing can provide valuable insights into the redistribution and accumulation of migrants on the ground, which cannot be measured by weather radars directly \citep{nussbaumer2024turnover}. Combining this information with citizen science data \citep{sullivan2014ebird} could help predict changes in spatio-temporal species distributions as well as local species compositions.

For simplicity, we have used a rather simple spatial interpolation to define the mapping from sparse radar observations to grid cells during encoding and model initialization. Augmenting this by a learned correction term or replacing it by a graph neural network layer similar to  \citet{vanderlinden2023learned} could reduce information loss and thus further improve prediction quality.
Alternatively, the FluxRGNN+ approach could be combined directly with some form of data assimilation (or state estimation) to obtain the best possible estimates of previous system states while accounting for uncertainties due to both imperfect measurements and model errors, as it is common in the Earth sciences \citep{reichle2008data, bauer2015quiet}.

Generally, hybrid forecast models like FluxRGNN+ have broad potential for ecological applications where complex spatio-temporal dynamics are sparsely observed and only incomplete theoretical models are available.
In such settings, black-box deep learning models designed for data-rich contexts and purely numerical simulations relying on fully specified models fall short.
Instead, hybrid models can explore the continuum between these two approaches, providing high-quality predictions and strong extrapolation capabilities, while offering opportunities to advance existing scientific knowledge and theories.

\newpage

\bibliography{references}
\bibliographystyle{icml2024}

\newpage

\appendix
\onecolumn
\section*{Appendices}

Appendix \ref{ap:data} provides additional information about the data used in our experiments. Appendix \ref{ap:models} discusses the FluxRGNN+ architecture and the considered baselines in more detail. Finally, we present additional results in Appendix \ref{ap:results}. 

\section{Data}\label{ap:data}

\subsection{Weather radar data}\label{ap:data:radar}

We use weather radar data downloaded from the NEXRAD archive \citep{ansari2018unlocking}, covering the autumn migration season (1 August to 15 November) of 2013-2021.
For each of the 143 radars, the vol2bird algorithm \citep{dokter2011bird} was applied to compute vertical profiles of bird densities and velocities based on measurements within a 5-35 km range around the radar. The settings of this algorithm, including filtering of meteorological signals, insects, ground clutter, and artifacts due to beam blockage, as well as velocity dealiasing, are the same as described in \citet{dokter2018seasonal}.
These profiles are further aggregated across altitudes, resulting in vertically integrated bird densities [birds/km\textsuperscript{2}] and average velocities [m/s], covering 3000 km above the radar antenna. 
Finally, bird density and velocity time series were resampled to 1-hour time intervals.

Additional manual filtering and data selection was performed to obtain high-quality ground truth data.
This involved (i) forcing bird densities and velocities during daytime to be zero, as most migration occurs during nighttime and daytime radar signals are likely to be contaminated with insect echoes, (ii) excluding measurements that are exactly zero due to high filter activity, and (iii) manual checks for artifacts or severe precipitation in radar scans with bird densities above 1165 birds /km\textsuperscript{2}.

To stabilize training, final bird densities are scaled by a factor of $0.001$, whereas velocities (and other spatial features of the tessellation like cell faces, areas and distances) are scaled by a factor of $0.01$.

\subsection{Time-varying input features}

% To capture the effects of dynamically changing environmental conditions on migration behaviors, we consider a set of time-varying features $\mathbf{u}^{(k)}_i \forall C_i\in\mathcal{C}$ capturing the state of the atmosphere as well as the current  time in the daily and annual cycles.  

\paragraph{Atmospheric reanalysis}

We downloaded data on zonal and meridional wind components, temperature, specific humidity, total cloud cover, total precipitation, surface pressure and mean sea level pressure from the ERA5 reanalysis database \citep{hersbach2020era5} with a time resolution of 1 hour and a spatial resolution of 0.5\textdegree.
For variables varying with altitude, we extracted data at three different model levels (135, 128, 115) matching the 10, 50 and 90\% quantiles of the average vertical distribution of bird densities across all years and radars. Model levels are a combination of pressure levels and above ground elevation, and therefore smoothly follow the topography without bearing the risk of extrapolating below the Earth surface. For the ICAO Standard Atmosphere Model, the considered levels correspond roughly to 54, 288, and 1329 meters above ground.
To match the spatial resolution to the model tessellation $\mathcal{C}$, all variables were aggregated into a single value per cell and time point, using the average of all data points falling within a cell.
These values are then normalized to $\left[-1, 1\right]$.

\paragraph{Daily and annual cycles}
To capture more general daily and seasonal trends, information on the daily and annual cycles are included in the form of the local solar position (determined for the cell center), its derivative (i.e. 1h change in solar position), a binary variable indicating if it is day or night, two binary variables marking civil dusk and dawn, as well as the day of the year.
The day of the year is normalized to $\left[0, 1\right]$, whereas solar positions are normalized to $\left[-1, 1\right]$.

Together, atmospheric variables and time-related variables form the vector of environmental conditions $\mathbf{u}_i^{(k)}$ for each cell $C_i$ and time point $t_k$ respectively.

\subsection{Time-invariant input features}

In addition to time-varying features, we also consider time-invariant properties of the tessellation. This includes information on land cover types as well as geographic locations of cell centers.

\paragraph{Land cover types}
Information on landscape characteristics stems from the National Land Cover Database (NLCD) \citep{yang2018new}, which classifies land cover types across North America at a 30m resolution. For each cell, we construct a vector $\mathbf{c}_i \in \left[0, 1\right]^{16}$, containing the proportions of that cell being covered by the 16 different land cover classes.

\paragraph{Geographic locations}
The location of each cell center is encoded using a sinusoidal embedding of longitude and latitude, resulting in a 4-dimensional vector $\mathbf{p}_i = \left[\sin(\text{lon}), \cos(\text{lon}), \sin(\text{lat}), \cos(\text{lat})\right] \ \in\left[-1, 1\right]^4$.

\section{Model details}\label{ap:models}

\subsection{FluxRGNN+}

\subsubsection{Encoder-decoder backbone}

Following \citet{lippert2022learning}, we use a 1-layer LSTM \cite{hochreiter1997long} to define $\text{RNN}_{\text{enc}}$ and $\text{RNN}_{\text{dec}}$ respectively.
At each time step, the inputs to these LSTMs consist of cell-level radar measurements $\Tilde{\mathbf{Y}}_{\mathcal{C}}^{(k)}$ (for the encoder) or predictions $\boldsymbol{\rho}^{(k)}_{\mathcal{C}}$ (for the decoder), as well as time-varying and time-invariant input features.
As the spatial context of a cell (e.g. proximity to the coast or mountain ranges) can affect the responses of birds to atmospheric conditions, we pass the land cover and location embedding vectors through a 2-layer graph attention network (GAT) \citep{velivckovic2018graph} before feeding it to the LSTMs.
In addition, we feed the final encoder embedding $\mathbf{Z}^{(0)}$ to the decoder at every time step $k>0$ to provide temporal context.

\subsubsection{Flux model}

As discussed in Section \ref{sec:methods:interpretability}, we explicitly predict average cell velocities and compute cell-to-cell fluxes $F_{i\to j}$ based on a numerical FVM scheme.
Similar to the original FluxRGNN parameterization, velocities $\mathbf{v}_i^{(k)}$ are the output of a space and time-invariant multi-layer perceptron (MLP), taking the corresponding decoder hidden state $\mathbf{z}^{(k)}_i$ together with relevant environmental variables $\mathbf{u}_i^{(k)}$ and the location embedding ${\mathbf{p}_i}$ as input:
\begin{equation}
    \mathbf{v}_i^{(k)} = \text{MLP}_{v}\left(\mathbf{z}^{(k)}_i, \mathbf{u}^{(k)}_i, {\mathbf{p}}_i \right)
\end{equation}
As velocities are unconstrained, no output non-linearity is applied.

Given the predicted velocities and current (predicted or initial) bird densities, we use a simple upwind scheme to compute numerical fluxes
\begin{equation}
    F_{i\to j}^{(k\to k+1)} = |f_{ij}| \left(a^+\rho_i^{(k)} + a^-\rho_j^{(k)}\right)
\end{equation}
with 
\begin{align}
    a^+ &= \max(0, \mathbf{n}_{ij}^T\mathbf{v}_{ij}^{(k)}) \\
    a^- &= \min(0, \mathbf{n}_{ij}^T\mathbf{v}_{ij}^{(k)}).
\end{align}
Here, $\mathbf{v}_{ij}^{(k)} = \frac{1}{2}(\mathbf{v}_i^{(k)} + \mathbf{v}_j^{(k)})$ approximates the velocity across face $f_{ij}$ between $C_i$ and $C_j$, and $\mathbf{n}_{ij}$ denotes the unit vector normal to this face and pointing outward of $C_i$.
Importantly, it is $F_{i\to j}^{(k\to k+1)}=-F_{j\to i}^{(k\to k+1)}$, and thus local mass conservation is ensured.

Note that this first-order approximation may not yield adequate results in modern fluid dynamics simulations, but it is sufficient for our purposes. Moreover, the neural networks parameterizing velocities and source/sink terms can in principle learn to counteract small numerical errors in the flux computation.

\subsubsection{Source/sink model}

For the source/sink term, we follow \citet{lippert2022learning} and parameterize
\begin{align}
    s_i^{(k\to k+1)} &= \gamma^{(k\to k+1)} - \delta_i^{(k\to k+1)}\rho_i^{(k)},
\end{align}
with learnable $0 \leq \delta_i^{(k\to k+1)} \leq 1$ and $\gamma^{(k\to k+1)}\geq 0$ representing the fraction of birds leaving cell $C_i$ to land, and the average increase in bird density in cell $C_i$ due to take-off, respectively. 
Note that ${s}_i^{(k\to k+1)}$ can, to a certain extent, act as a learnable correction term accounting for numerical errors in the flux estimation.

Again, $\delta_i^{(k\to k+1)}$ and $\gamma^{(k\to k+1)}$ are predicted as a function of the encoder hidden state, environmental conditions, and the cell location embedding, using a shared space and time-invariant MLP
\begin{equation}
    {\delta}_i^{(k\to k+1)}, {\gamma}_i^{(k\to k+1)} = \text{MLP}_{s}\left(\mathbf{z}^{(k)}_i, \mathbf{u}^{(k)}_i, {\mathbf{p}}_i \right).
\end{equation}
To ensure that $\delta_i^{(k\to k+1)}$ and $\alpha^{(k\to k+1)}$ are in the correct ranges, we use the following output non-linearities:
\begin{align}
    \delta_i^{(k\to k+1)} = \left(\tanh\left(\cdot\right)\right)^2 \quad \text{and} \quad
    \gamma_i^{(k\to k+1)} = \left(\cdot\right)^2.
\end{align}
Note that this setup encourages initial (random) predicted source and sink terms to be close to zero, while promoting accurate predictions of extreme migration events.

\subsubsection{Boundary conditions}\label{ap:models:FluxRGNN+:boundary}

% Migratory movements are likely to extend beyond the considered spatial domain $\Omega$. That means that fluxes across the boundary cannot be assumed to be zero. Instead, 
Assuming a smooth enough process, bird densities and hidden states at boundary cells $B_i\in\mathcal{B}\subset\mathcal{C}$ are defined based on the average of neighboring cells $C_j\in\mathcal{N}_{\mathcal{C}}(i)\setminus\mathcal{B}$. This can be interpreted as an approximate Neumann boundary condition $\frac{\partial\rho}{\partial \mathbf{n}}(\mathbf{x})=0 \ \forall\mathbf{x}\in\partial\Omega$ where $\mathbf{n}$ denotes the normal to the domain boundary $\partial\Omega$.

\subsubsection{Loss function}
The bird density and velocity components of the FluxRGNN+ loss function (see section \ref{sec:methods:interpretability}) are defined as
\begin{align}
        \mathcal{L}_{\rho}(\boldsymbol{\rho}_{\mathcal{R}}^{(k)}, f_{\mathcal{C}\to\mathcal{R}}\left(\boldsymbol{\rho}_{\mathcal{C}}^{(k)}\right) &= \frac{1}{M} \sum_{m=1}^M \lVert\boldsymbol{\rho}^{(k)} - f_{\mathcal{C}\to\mathcal{R}}\left({\boldsymbol{\rho}}_{\mathcal{C}}^{(k)}\right)\rVert_2^2 \\
        \mathcal{L}_{v}(\mathbf{v}_{\mathcal{R}}^{(k)}, f_{\mathcal{C}\to\mathcal{R}}\left(\mathbf{v}_{\mathcal{C}}^{(k)}\right) &= \frac{1}{M} \sum_{m=1}^M \lVert\mathbf{v}_{x, m}^{(k)} - f_{\mathcal{C}\to\mathcal{R}}\left({\mathbf{v}}_{x,\mathcal{C}}^{(k)}\right)\rVert_2^2 + \lVert\mathbf{v}_{y, m}^{(k)} - f_{\mathcal{C}\to\mathcal{R}}\left({\mathbf{v}}_{y,\mathcal{C}}^{(k)}\right)\rVert_2^2
\end{align}
where $\mathbf{v}_x$ and $\mathbf{v}_y$ are the East-West and North-South components of velocity vectors.

\subsubsection{Training and evaluation}

We train FluxRGNN+ on 72-hour sequences, where the first 24 hours are passed to the encoder network and the remaining 48 hours are used for training. Training sequences are constructed using a sliding window, resulting in 2424 partially overlapping sequences per season.
Starting from a maximum horizon of $K=2$ hours, we linearly increase $K$ every 10 epochs until $K=48$ or the maximum number of epochs is reached.

For model evaluation, we also use a 24-hour context, but generate forecasts up to 72 hours into the future. To mimic an operational forecast setting, we only consider sequences starting at 13:00 (Eastern Time), which means that 72h-forecasts cover three complete migration nights. If not indicated otherwise, the reported evaluation metrics are averaged across the full forecast length, excluding daytime hours.

For both training and evaluation, sequences with more than $10\%$ missing data were excluded.

\subsection{FluxRGNN (Voronoi tessellation)}
We compare our modified model to the original FluxRGNN model operating on the Voronoi tessellation of radar locations. To ensure a fair comparison, we match the experimental setup and the number of parameters as much as possible. 
That means, we consider a FluxRGNN setup that differs only in two major points from our proposed model: (i) the underlying tessellation used for spatial discretization of the movement process, and therefore the radar-to-cell and cell-to-radar mappings, and (ii) the neural parameterization of flux terms.

\subsubsection{Voronoi tessellation}
We construct the Voronoi tessellation based on the locations of the 143 NEXRAD radars and 60 additional dummy radars which are positioned in regular intervals along the boundary of the observed domain (defined based on a 450km buffer around observed radar locations).
The cells corresponding to these dummy radars form the (unobserved) boundary $\mathcal{B}$ to which the boundary extrapolation (see section \ref{ap:models:FluxRGNN+:boundary}) is applied.

%TODO: something about the distribution of cell sizes and distances between radars?

\subsubsection{Flux parameterization}
In contrast to our velocity-based flux model, \citet{lippert2022learning} parameterize fluxes as
\begin{align}
    F_{i\to j}^{(k\to k+1)} &= \mathbf{A}^{(k\to k+1)}_{ji}V_i\bar{\rho}_i^{(k)}  - \mathbf{A}^{(k\to k+1)}_{ij}V_j\bar{\rho}_j^{(k)},
\end{align}
with learnable $\mathbf{A}^{(k\to k+1)}_{ij}\in\left[0, 1\right]$ representing mass flow rates, i.e. the fraction of birds moving from cell $C_j$ to cell $C_i$.
This construction avoids the explicit parameterization of the underlying velocity field $\mathbf{v}$, and thus circumvents the need for approximate FVM schemes.
However, it prevents additional supervision in cases where velocity measurements are available.

The flow rates are the output of a space and time-invariant MLP, taking as input past conditions in the upstream cell ($\mathbf{z}_j^{(k)}$, and $\mathbf{u}_j^{(k)}$), current conditions in the downstream cell ($\mathbf{u}_i^{(k+1)}$), as well as edge features $\mathbf{e}_{ij}$ containing information about cell distances, directions, and face lengths:
\begin{equation}
    \mathbf{A}_{ij}^{(k\to k+1)} = \text{MLP}_{A}\left(\mathbf{z}^{(k)}_j, \mathbf{u}^{(k)}_j, \mathbf{u}^{(k+1)}_i, {\mathbf{e}}_{ij} \right)
\end{equation}

\subsection{XGBoost baseline}

Following \citet{vandoren2018continental}, we used XGBoost to predict cube-root-transformed bird densities based on local environmental conditions. 
Input features consist of the same time-varying variables as used for FluxRGNN and FluxRGNN+, together with sinusoidal embeddings of radar locations.

\subsection{Model hyperparameters}

For FluxRGNN+ and XGBoost, we perform hyperparameter sweeps where we randomly select 5 subsets each consisting of 15 radars which are held out during training and serve as an independent validation set to evaluate model performance for unseen locations. The hyperparameter setting resulting in the best average performance across these 5 cross-validation folds is chosen for further experiments.
The considered hyperparameters and final settings are summarized in Tables \ref{tab:hyperparameter_search_fluxrgnn} and \ref{tab:hyperparameter_search_xgboost}.
Hyperparameters for the FluxRGNN baseline are chosen to match to ones used for FluxRGNN+.

\begin{table}[htb]
    \caption{Considered hyperparameter space and final settings for FluxRGNN+ and FluxRGNN. Note that the number of hidden channels was not calibrated independently for each model component, but the same value was used for $\text{MLP}_v, \text{MLP}_s, \text{RNN}_{\text{enc}}$ and $\text{RNN}_{\text{dec}}$.\newline}
    \label{tab:hyperparameter_search_fluxrgnn}
    \centering
    % \scriptsize
    \vspace{2ex}
    \begin{tabular}{llcc}
        \toprule
        component & hyperparameter & search space & final setting \\
        \midrule
        \multirow{2}{*}{$\text{MLP}_{v}$} & \# layers & \{1, 2\} & 1 \\
        & \# hidden channels & \{32, 64, 128\} & 128 \\
        \midrule
        \multirow{2}{*}{$\text{MLP}_{s}$} & \# layers & \{1, 2\} & 1 \\
        & \# hidden channels & \{32, 64, 128\} & 128 \\
        \midrule
        \multirow{2}{*}{GAT} & \# layers & \{1, 2, 3\} & 2 \\
        & \# hidden channels & \{16, 32, 64\} & 32 \\
        \midrule
        $\text{RNN}_{\text{enc}}$ & \# hidden channels & \{32, 64, 128\} & 128 \\
        \midrule
        $\text{RNN}_{\text{dec}}$ & \# hidden channels & \{32, 64, 128\} & 128 \\
        \midrule
        \multirow{6}{*}{General settings} & learning rate & $\{3\times 10^{-6}, 10^{-5}, 3\times 10^{-5}, 10^{-4}, 3\times 10^{-4}\}$  & $10^{-4}$ \\
        &  dropout & \{0\%, 10\%, 25\%\} & 10\%\\
        &  activation function & -- & ReLU \\
        &  optimizer & -- & Adam\\
        &  batch size & -- & 32 \\
        &  epochs & -- & 500\\
        \bottomrule
    \end{tabular}
\end{table}

\begin{table}[htb]
    \caption{Considered hyperparameter space and final settings for XGBoost. \newline}
    \label{tab:hyperparameter_search_xgboost}
    \centering
    \vspace{2ex}
    \begin{tabular}{lcc}
        \toprule
        hyperparameter & search space & final setting \\
        \midrule
        num estimators & -- & 500 \\
        max depth & \{5, 10, 15, 20\} & 15 \\
        learning rate & \{0.01, 0.05, 0.1\} & 0.05\\
        min child weight & \{1, 3, 5\} & 3 \\
        subsample & \{0.7, 1.0\} & 0.7 \\
        gamma & \{0, 1, 10\} & 0 \\
        \bottomrule
    \end{tabular}
\end{table}

\section{Additional results}\label{ap:results}

\subsection{Predictive performance at observed and unobserved locations}

\begin{figure*}[htb]
    \centering
    \includegraphics[width=0.8\textwidth]{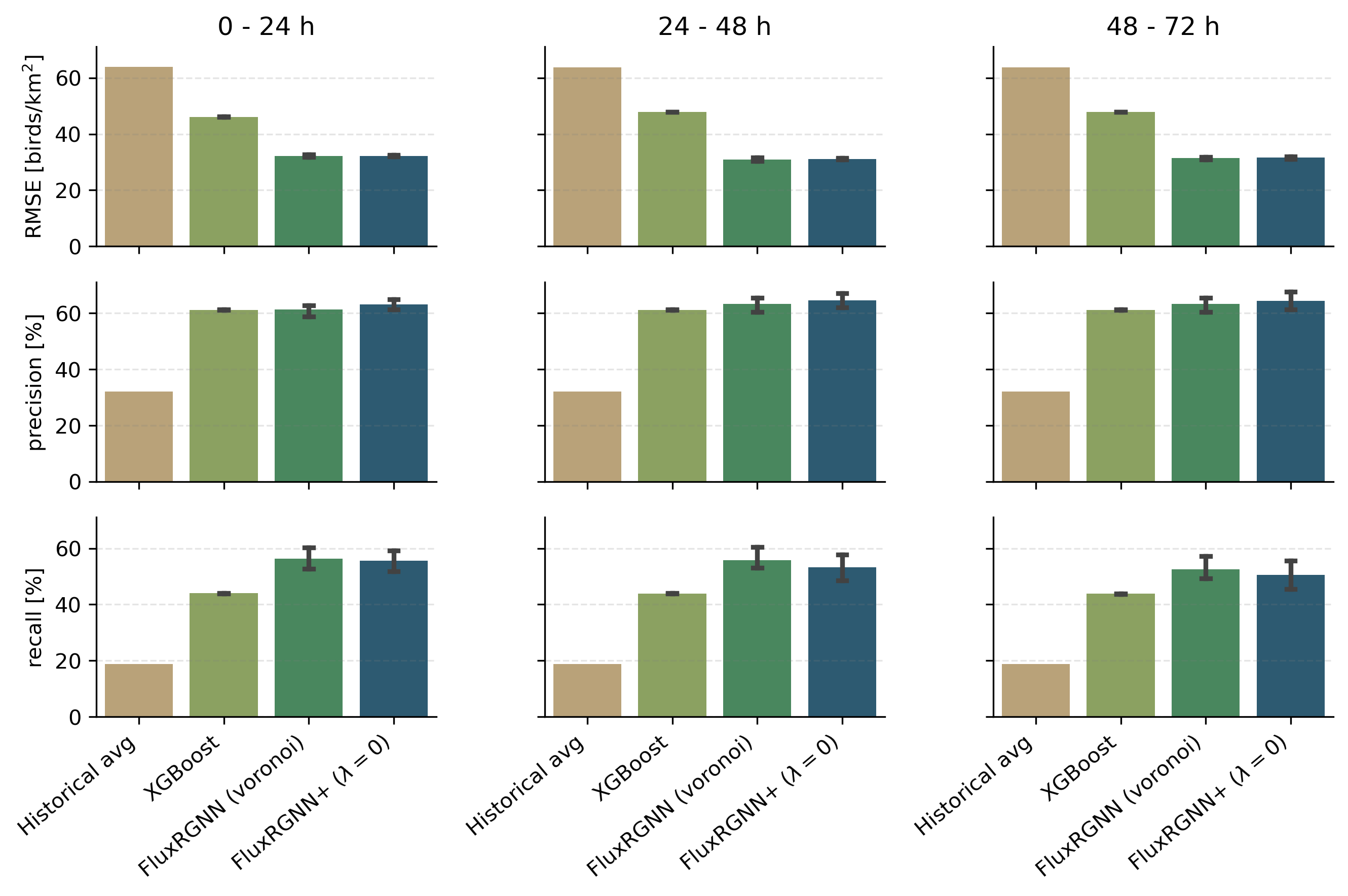}
    \caption{Evaluation of bird density predictions, separated by forecasting horizons. All metrics are reported as mean $\pm$ std across 5 different random seeds.}
    \label{ap:fig:all_radars_rmse_per_night}
\end{figure*}

Table \ref{tab:all_radars_performance} in the main text summarizes the predictive performance at observed radar locations across all time steps and radars. Here, we provide more detailed results where we separate each forecast into 24h-bins and compute evaluation metrics for each bin respectively. 
We find that both FluxRGNN and FluxRGNN+ continue to outperform local baseline models even for horizons beyond those considered during training, i.e. 48-72h (see Figure \ref{ap:fig:all_radars_rmse_per_night}). This indicates that the hybrid approach results in robust predictions and minimal error propagation.

\begin{figure*}[htb]
    \centering
    \includegraphics[width=0.8\textwidth]{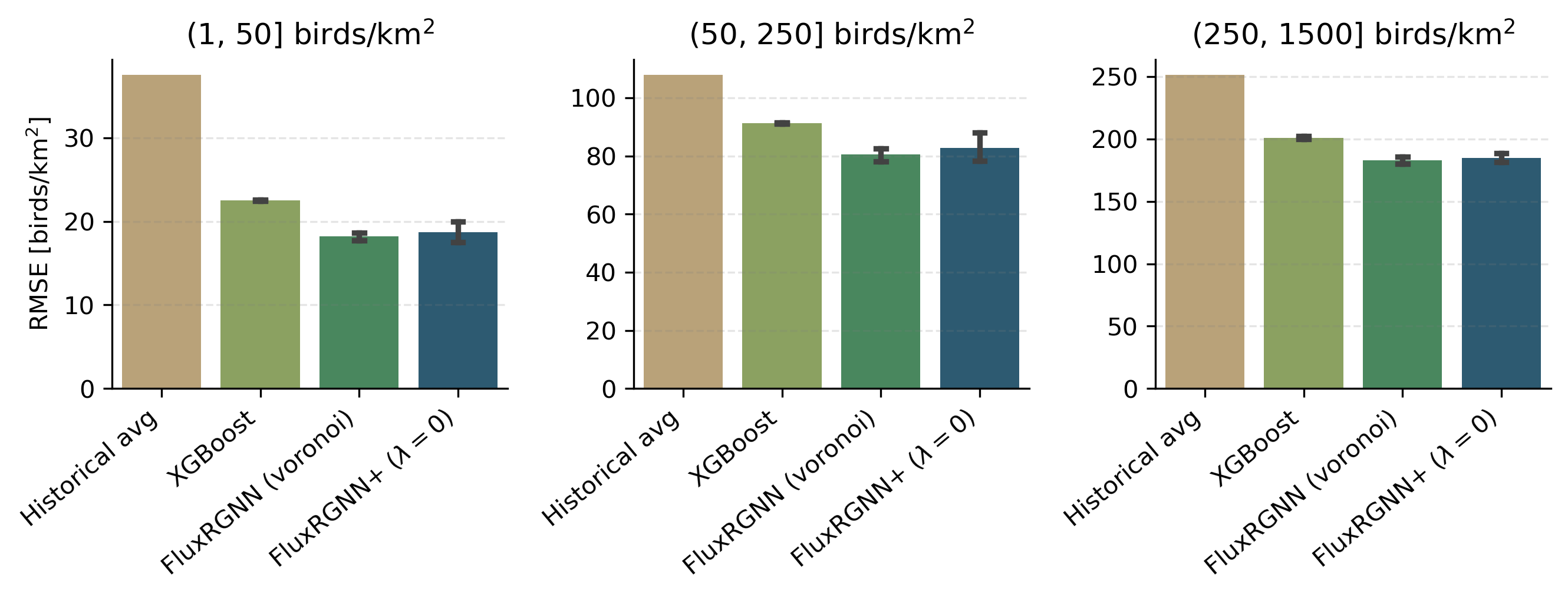}
    \caption{Evaluation of bird density predictions, separated by ground truth bird densities. All metrics are reported as mean $\pm$ std across 5 different random seeds.}
    \label{ap:fig:all_radars_rmse_3bins}
\end{figure*}

\begin{figure*}[htb]
    \centering
    \includegraphics[width=0.8\textwidth]{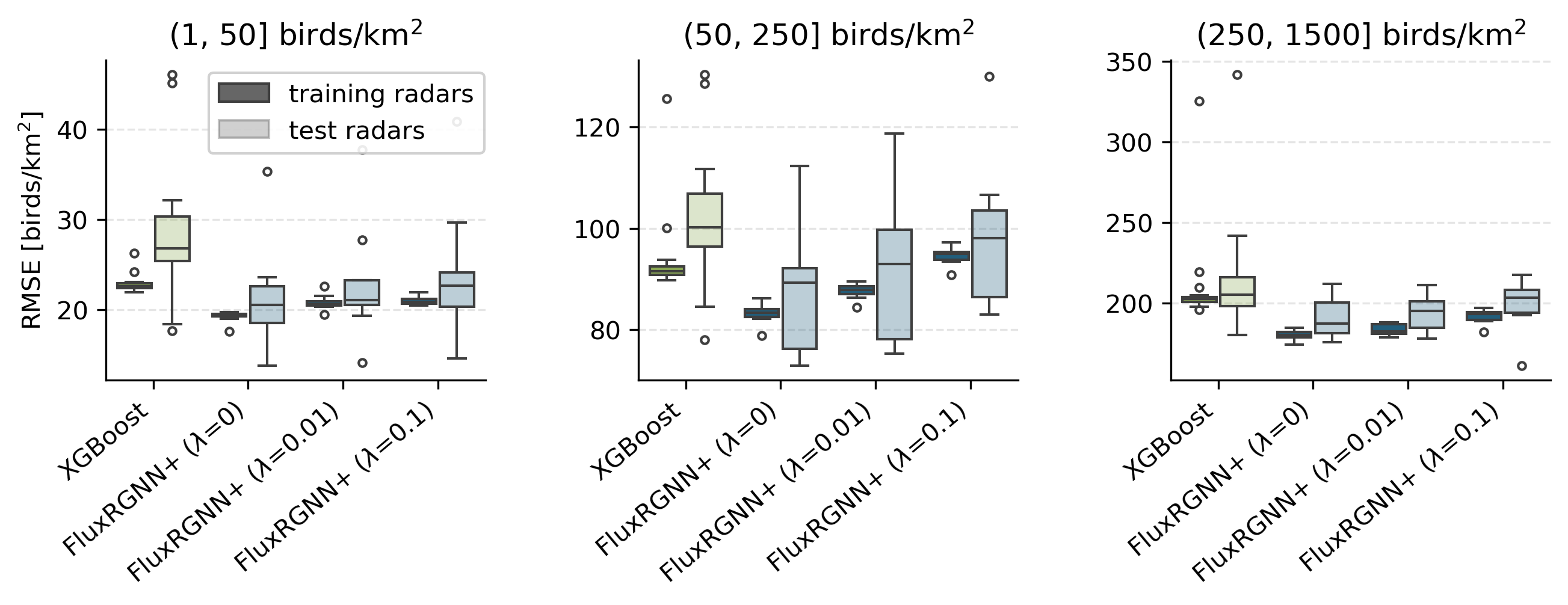}
    \caption{Spatial cross-validation of bird density predictions, separated by ground truth bird densities. Box plots show the variability of evaluation metrics across 10 cross-validation folds, where different subsets of radars were held out during training.}
    \label{ap:fig:cv_rmse_3bins}
\end{figure*}

Similarly, we separate predictions according to observed bird densities to gain insights into the quality of predictions during low, medium and high intensity migration events. As expected, Figures \ref{ap:fig:all_radars_rmse_3bins} and \ref{ap:fig:cv_rmse_3bins} shows that the RMSE for both observed and unobserved radar locations increases as ground truth bird densities increase. The relative performance of models remains very similar, however, with FluxRGNN and FluxRGNN+ being more accurate than local baselines for low, medium and high migration events respectively.

\begin{figure*}[htb]
    \centering
    \includegraphics[width=0.7\textwidth]{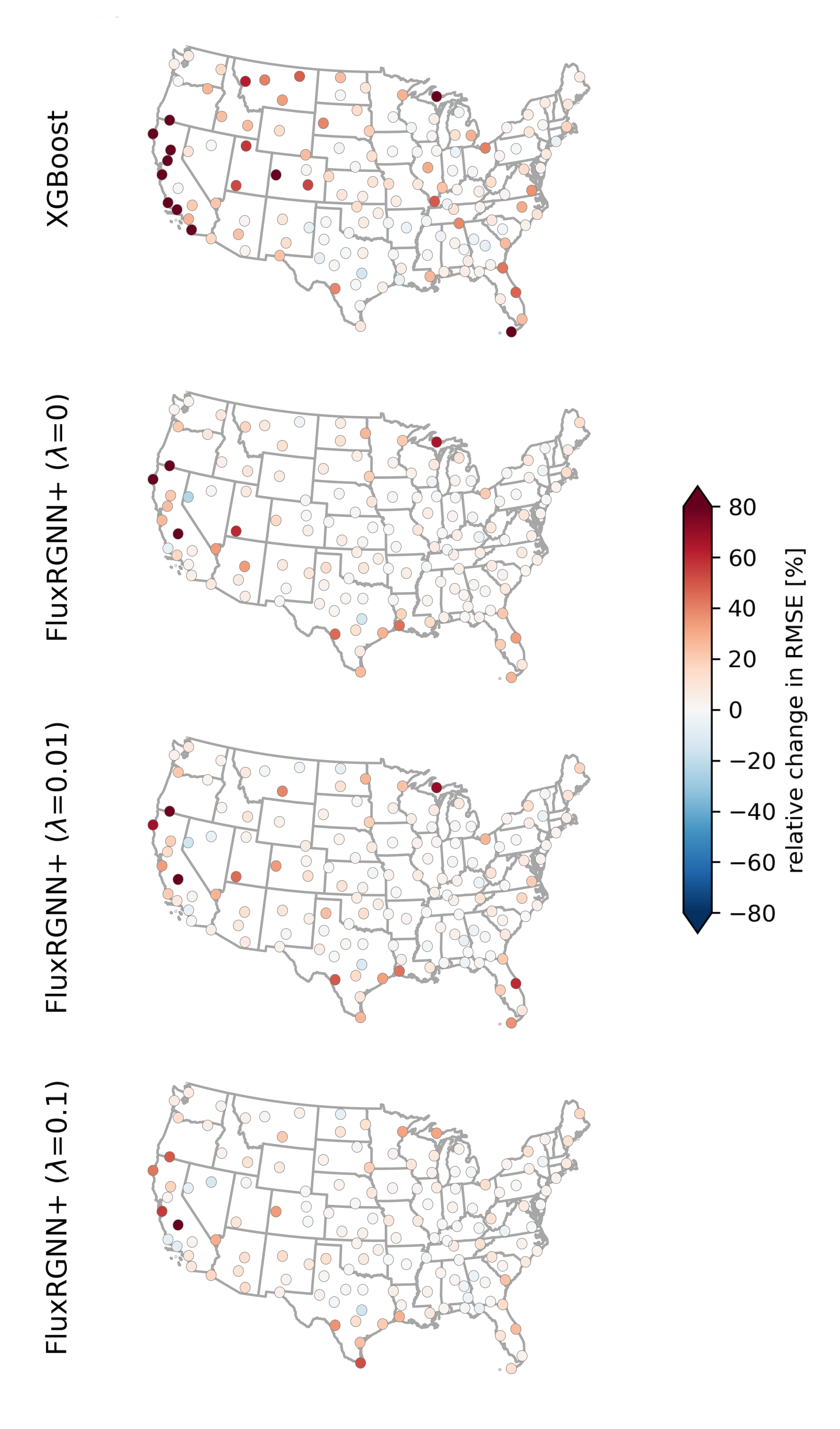}
    \caption{Radar-specific cross-validation of bird density predictions in terms of the relative change in RMSE, indicating the increase in prediction error of a radar due to missing observations.}
    \label{ap:fig:cv_rel_change_rmse}
\end{figure*}

In addition, we perform a radar-specific evaluation. For each radar, we compute both the average RMSE across folds where the radar is included during training and the RMSE for the fold where it is considered unobserved. We then compute the relative change in RMSE, indicating by how much the prediction error increases due to missing observations.
Figure \ref{ap:fig:cv_rel_change_rmse} shows the spatial distribution of this quantity for the different models. Clearly, XGBoost suffers the most from missing observations. All three FluxRGNN+ variants show very strong extrapolation capabilities, especially for radars away from the domain boundary.

\subsection{Example forecast}

\begin{figure*}[htb]
    \centering
    \includegraphics[width=\textwidth]{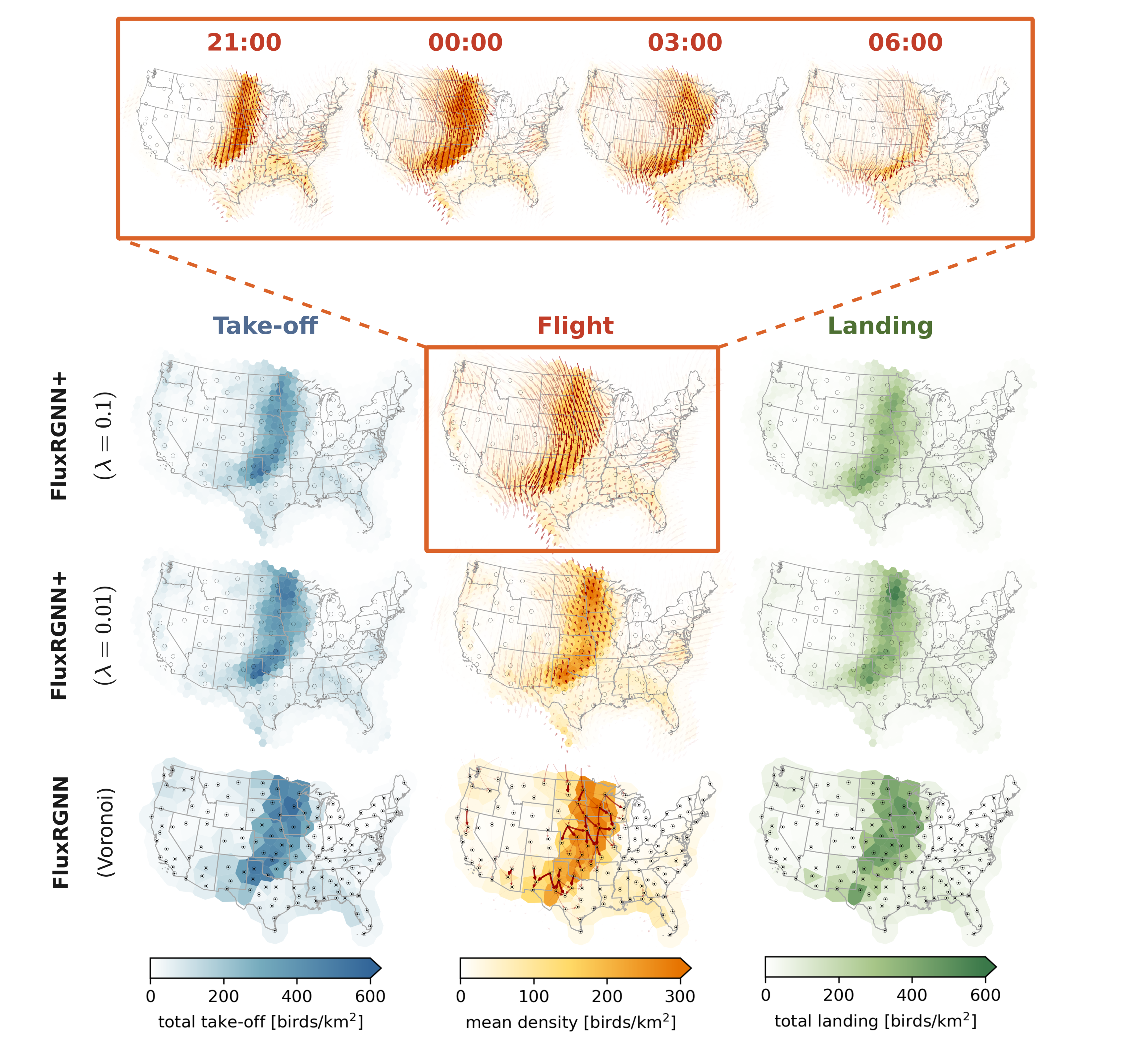}
    \caption{Example forecasts for a high intensity migration night (number 1 in Figure \ref{fig:example_forecast}) in September 2021 generated by three different models trained on years 2013-2020. To obtain predictions of total take-off and landing, we separate hours with positive and negative source/sink term and aggregate them respectively. Red arrows in areas with substantial migration indicate average velocities (for FluxRGNN+) and cell-to-cell fluxes (for FluxRGNN) respectively. Top: bird density snapshots at different time points throughout the night (Eastern Time Zone), predicted by FluxRGNN+ with $\lambda=0.1$.}
    \label{ap:fig:example_forecast_comparison}
\end{figure*}

Finally, we provide a visual comparison of predictions generated by the original FluxRGNN and our extended model using different settings for $\lambda$ (see Figure \ref{ap:fig:example_forecast_comparison}).
Overall, the predicted migration flows match well. FluxRGNN+ clearly provides much more detailed spatial patterns than the original FluxRGNN based on Voronoi cells. In line with findings from Figure \ref{fig:speed_dir_performance}, we see that FluxRGNN+ trained with $\lambda=0.01$ tends to underpredict flight speeds resulting in a rather unrealistic combination of high bird densities but little aerial movements.

\end{document}